\documentclass{article} % For LaTeX2e
\usepackage{times}

\usepackage{hyperref}
\usepackage{url}
\usepackage{graphicx}
\usepackage{subcaption}
\usepackage{booktabs}

\title{High-frequency crowd insights for \\ public safety and congestion control}

\author{Karthik.~Nandakumar, Sebastien Blandin, Laura Wynter \\
IBM Research\\
Singapore\\
\texttt{\{nkarthik, sblandin, lwynter\}@sg.ibm.com} \\
}

\begin{document}

\maketitle

\begin{abstract}
We present results from several projects aimed at enabling the real-time understanding of crowds and their behaviour  in the built environment. We make use of CCTV video cameras that are ubiquitous throughout the developed and developing world and as such are able to play the role of a reliable sensing mechanism. We outline the novel methods developed for our crowd insights engine, and illustrate examples of its use in different contexts in the urban landscape. Applications of the technology range from maintaining security  in public spaces to quantifying the adequacy of public transport level of service.
\end{abstract}
%~~~~~~~~~~~~~~~~~~~~~~~~~~~~~~~~~~~~~~~~~~~~~~~~~~~~~~~~~~~~~~
\begin{section}{Introduction}
Increasing urbanization in the developing and developed world puts significant pressure on public infrastructure. High urban population leads to large crowds in public spaces which in turn can lead to a degradation of efficiency and public safety, if not well managed. The management of crowds at events and at critical infrastructures such as airports and public transport facilities is thus an important component in sustainable urban planning, especially under increasing demand pressure.

As of 2018, $33$ cities in the world have population exceeding $10$ million. According to~\cite{UNhabitat}, this number is expected to grow to $43$ in 2030. This urban population growth affects on a daily basis the public transport infrastructure. In Singapore, for example, average daily ridership increased from $5.01$ million in 2002 to $7.67$ million in 2014, see~\cite{lta}. In Hong-Kong, the monthly patronage increased from $66.07$ million in January 2000 to $148.22$ million in January 2019, see~\cite{mtr}.  Large scale gatherings and events are also affected, seeing increasing crowds and increasing chances of crowd-related incidents. An extreme example occurred in 2015 
 during the annual Muslim pilgrimage near the holy city of Mecca, in Saudi Arabia, when thousands of pedestrians were trampled.

The challenges of crowd monitoring and management in the above settings are significant,~\cite{faster2019}. There is a dramatic variation in usage,  requiring fast feedback during times of large crowds and incidents. Crowding events must be  precisely estimated in time, space, and magnitude as well as their nature (calm vs. angry), in order to support the following feedback mechanisms; (i) information to individuals within and outside the crowd, (ii) local prescriptive enforcement via ground personnel, (iii) regional adjustment of resources to manage the event. As the risk of incidents increases with the density of the crowd, along with crowd-level analysis, the precursors of such incidents are required so as to enable rapid mitigation.

This requirement on precision in both the nature of the behavior and the time-frame of changes to those parameters means that many technologies such as GPS-based crowd analysis are less effective than video, the latter enabling analysis of the full population rather than an unknown subsample, as well as important features such as movement and emotion. Much of  video analysis however aims  to account for each person individually; however, in crowded spaces, bodies and faces overlap.

Many of the requirements and challenges  were detailed two decades ago in the classic reference~\cite{davies1995crowd}. \cite{helbing2000simulating} studied crowd-related disasters and emphasized the need for monitoring of crowd behavior in densely-populated spaces. Mathematical aspects of crowd modeling can be found in~\cite{cristiani2014multiscale}.

We provide details of our system that both accurately estimates  crowd flow and density and detects events of interest in the crowds, using existing CCTV cameras. Our approach has the following advantages; (i) the infrastructure to sense, collect and retrieve data  exists  across cities in both the developing and developed world, and (ii)  accuracy is  limited mainly by the algorithm, rather than the infrastructure, and can be improved as computing power increases, unlike geo-localized signals such as GPS or indoor positioning technologies. 

% The challenges of crowd sensing and management on train platforms are exemplified in Figure~\ref{fig:intro}.
% \begin{figure}[htb!]
% \centering
% \begin{subfigure}{.40\columnwidth}
%     \centering
%     \includegraphics[width=\columnwidth]{img/otp.png}
% \end{subfigure}
% \begin{subfigure}{.3\columnwidth}
%     \centering
%     \includegraphics[width=\columnwidth]{img/dense_crowds.png}
% \end{subfigure}
% \begin{subfigure}{.25\columnwidth}
%     \centering
%     \includegraphics[width=\columnwidth]{img/image_inria.png}
% \end{subfigure}%
% \caption{Left: pendularity of crowd level in a typical urban train station. Middle: typical view from a CCTV camera on a train platform. Right: sample images from INRIA Person dataset.}
% \label{fig:intro}
% \end{figure}
\end{section}
%~~~~~~~~~~~~~~~~~~~~~~~~~~~~~~~~~~~~~~~~~~~~~~~~~~~~~~~~~~~~~~
\begin{section}{CCTV-based Crowd Monitoring}
The proposed  system consists of two main modules: (i) Crowd flow and density   estimation, and (ii) crowd anomaly detection, and uses both static features extracted from individual frames of the video and spatio-temporal features extracted from contiguous sequence of frames.

% \begin{figure}[htb!]
%     \begin{center}
%     \includegraphics[width=0.8\textwidth]{img/OverallSystemDiagram}
%     \caption{Overall schematic diagram of the proposed video-based crowd surveillance system.}\label{fig:SystemDiagram}
%     \end{center}
% \end{figure}

\begin{subsection}{Crowd Flow and Density   Estimation}
Our crowd level estimation module uses  adaptive fusion combining people detection with crowd-level image-feature based regression. The counting-by-detection approach scans the image frame for individual pedestrians and works well for sparsely crowded scenes (Figure~\ref{fig:SparseDenseCrowds} (a)), where there is  less occlusion. Regression-based counting directly estimates the people count in densely crowded scenes, where occlusion impedes detection of individuals (Figure~\ref{fig:SparseDenseCrowds} (b)). 

\begin{figure}[htb!]
\centering
\begin{subfigure}{.42\columnwidth}
    \centering
    \includegraphics[width=\columnwidth]{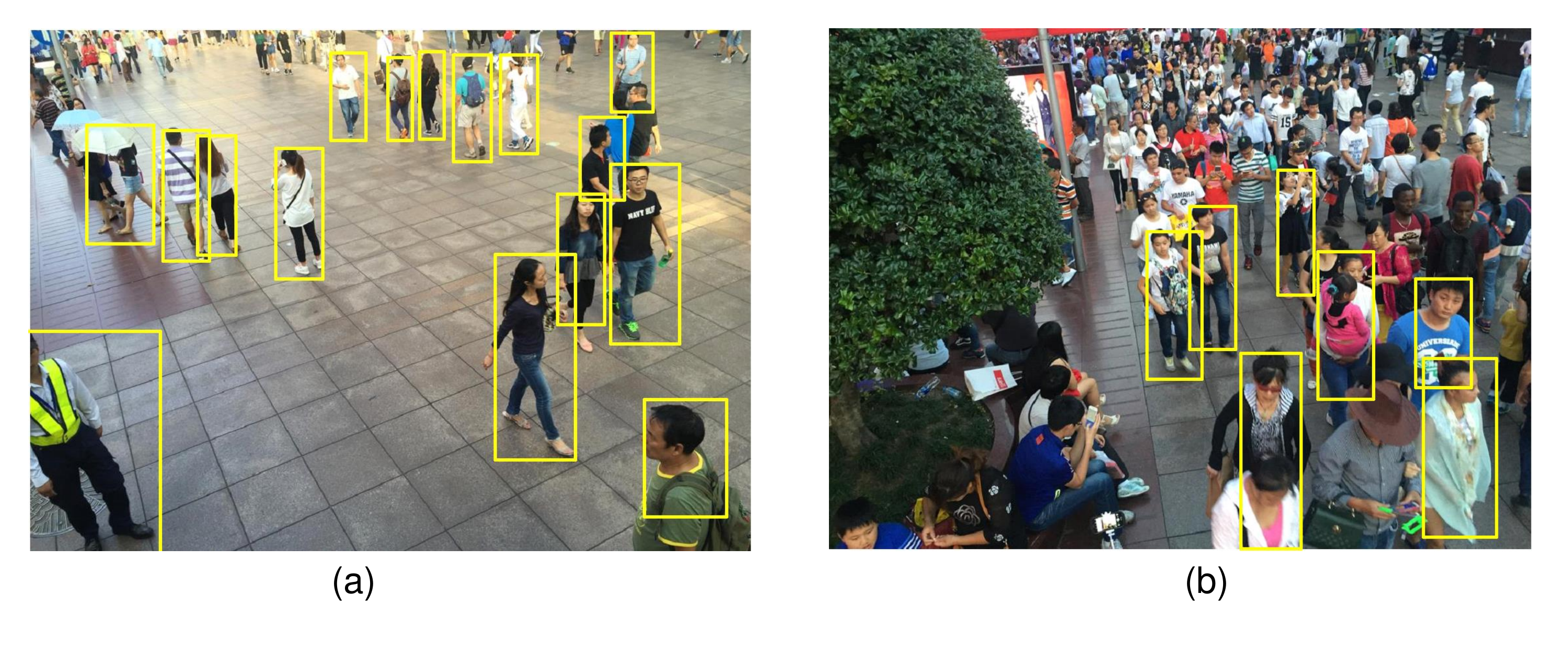}
\end{subfigure}
\begin{subfigure}{.48\columnwidth}
    \centering
    \includegraphics[width=\columnwidth]{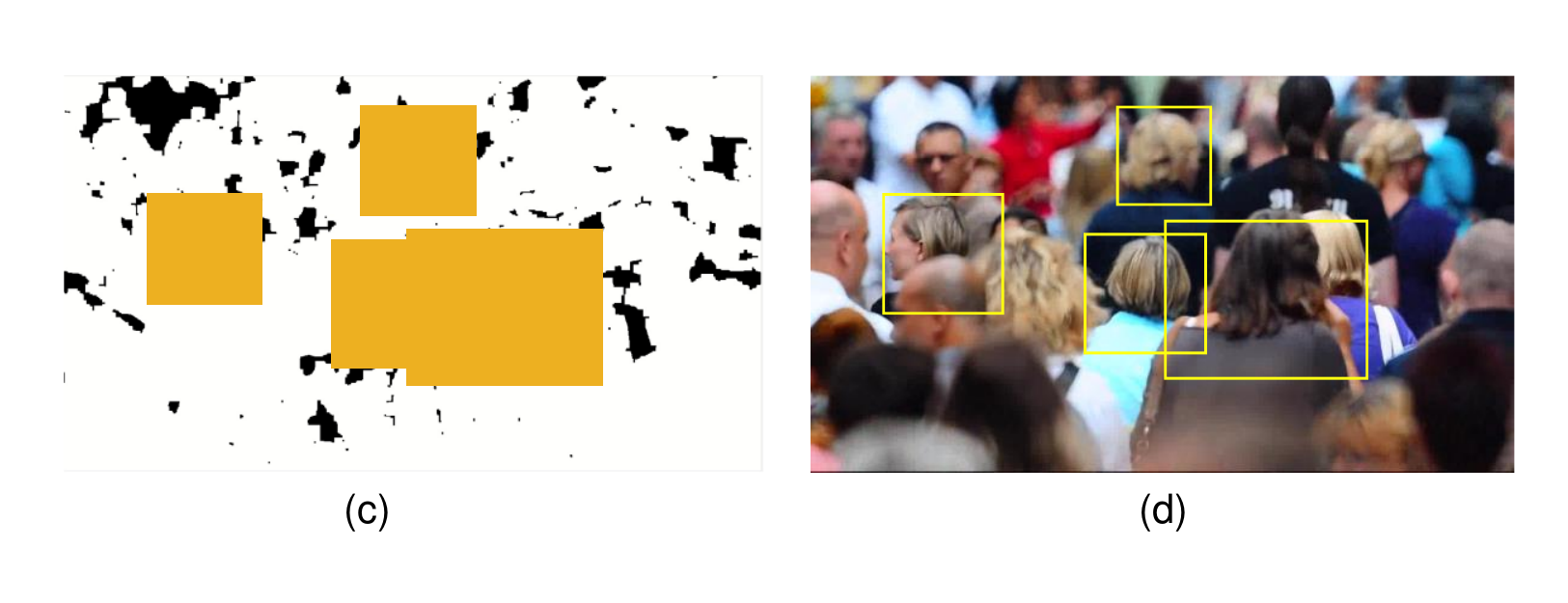}
\end{subfigure}
\caption{People counting by detection for (a) sparse crowd and (b) dense crowd. Occlusion leads to severe underestimation in (b). These images are extracted from the ShangaiTech dataset, \cite{zhang2016single}. Adaptive fusion  used for video-based people counting, (c) foreground segmentation and (d) pedestrian detection. The fusion model combines the people detections with the regression method after de-emphasizing the regions with pedestrian detection (marked by yellow boxes in (c)).}\label{fig:SparseDenseCrowds}
\end{figure}

\paragraph{People Detection and Flow:} The people detector, appropriate for low crowd levels,  detects features of individuals. Our algorithm is based on full-body, head and shoulder, and head characteristics. To improve robustness to occlusion, the algorithm extracts classical features such as histogram of Gradients (HoG), gradient magnitude, and texture, over multiple channels. Boosted decision trees are used as the classifier.
Extending the algorithm via  tracking of pedestrian trajectories, we obtain estimates of the people flow. Filtering is then performed to remove trajectories that are either very short or do not contain any strong directional movement. Finally, a two-class classifier is applied to categorize the remaining trajectories into incoming and outgoing pedestrian flows.

\paragraph{Regression-based Counting:} This component first extracts simple features (e.g., shape and texture) from the image, weighted using a perspective normalization. We divide the image into $12$ horizontal blocks and compute the ratio of foreground and edge pixels in each block. Thus, we obtain a $24$-dimensional feature vector, which is used as the input to a linear regression algorithm to obtain an estimate of the crowd density.
When  computational resources are sufficient, we use deep learning for feature extraction. Specifically, we use the basic AlexNet convolutional neural network~\cite{krizhevsky2012imagenet} and  the $4,096$-dimensional embedding obtained at the output of the last fully convolutional layer. A support vector regression method is used to obtain the crowd density estimate.

\paragraph{Adaptive Fusion:} Crowd levels at a specific location are highly variable and relying on a single approach leads to large errors when the conditions change. We thus developed an adaptive fusion scheme that robustly combines  the above approaches to improve the accuracy of crowd density estimates across all conditions. An illustration is shown in Figures~\ref{fig:SparseDenseCrowds} (c) and (d). The key element that leads to its success  is to extract features required for crowd-based regression only from regions where there are no pedestrian detections. As such, the method automatically assigns a higher importance to the detection-based count in sparsely crowded scenes and to the regression-based count in densely crowded scenes. We set a stringent threshold to minimize the number of false detections.
\end{subsection}

\begin{subsection}{Crowd Incident Detection}
We consider  crowd incidents to be characterized by a rapid movement of people towards each other with abrupt directional changes. The following two flow features are thus crucial. In particular, in order to reduce false positives, we discard features directly extracted from individual trajectories.

\paragraph{Aggregate flow patterns:} Assuming that the background is stationary and motion is primarily restricted to the people in the scene, optical flow-based features characterize crowd movement patterns at a coarse level. Figure~\ref{fig:AggregateFlowFeatures} shows  examples of aggregate flow-based features extracted from three scenarios (a fight, a group of people dancing, and dense moving crowd without any incident). While it may be easy to filter out the third scenario, distinguishing between the first two scenarios based on this set of features is challenging  as both  exhibit rapid and abrupt motions. 

The image is  divided into many smaller blocks, histogram of the aggregate flow field is obtained for each block, and these histograms are concatenated to obtain the global feature vector, which is finally input to a Support Vector Machine (SVM) classifier to detect fight/violence.

\begin{figure}[htb!]
    \begin{center}
    \includegraphics[width=\textwidth]{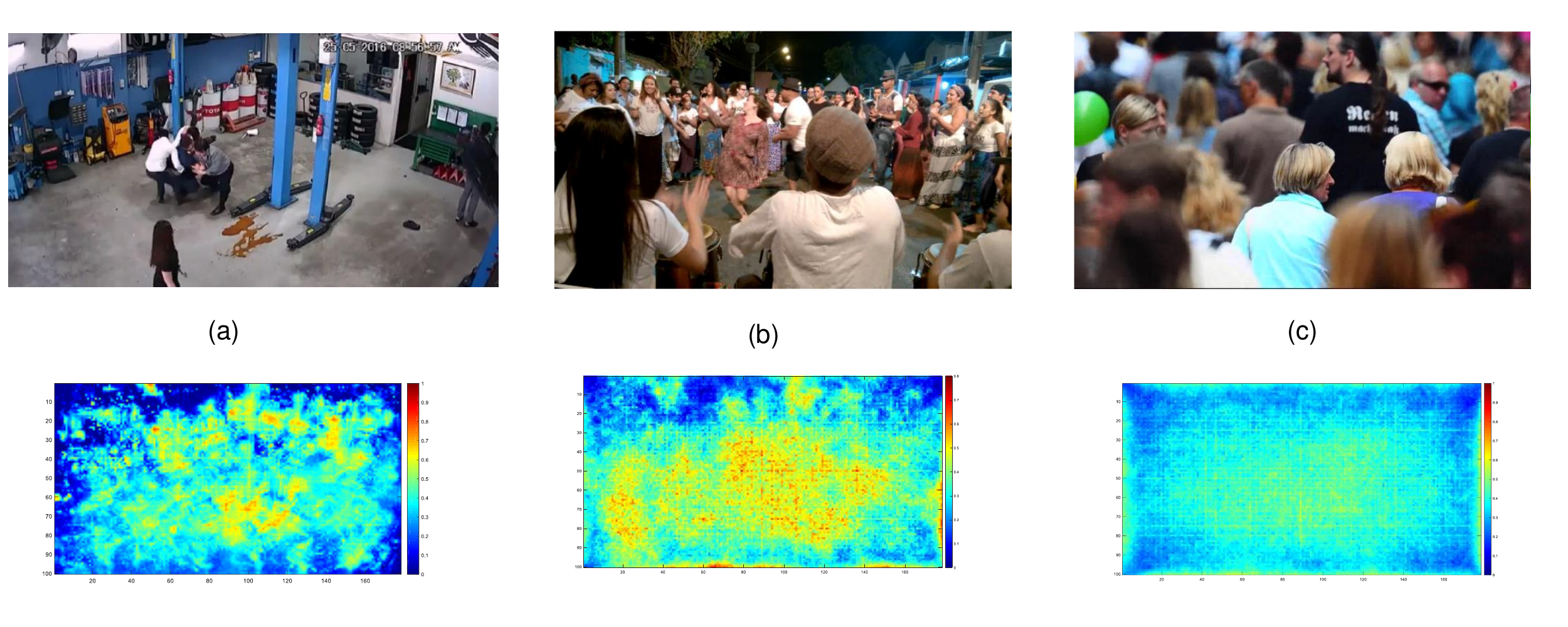}
    \caption{Flow-based features extracted from three different scenarios. (a) fight, (b) group of people dancing, and (c) dense moving crowd. Dark blue pixels indicate regions with low activity and the red pixels denote regions with rapid motion.}\label{fig:AggregateFlowFeatures}
    \end{center}
\end{figure}

\paragraph{Features extracted from selected interest points:} A subset of interest points is automatically selected (unlike the flow-estimator where the motion of all pixels is considered), and these specific interest points are tracked for short intervals (typically in the order of 0-5 seconds), generating what are known as tracklets. Spatio-temporal features are then extracted from tracklets. 

Three types of crowd interaction patterns have been proposed in~\cite{shao2015deeply}. While collectiveness represents the degree to which individuals move together, stability indicates the degree of change in the topological crowd structure, and conflict occurs when individuals move towards different directions. Collectiveness and conflict patterns are especially useful for abnormal crowd behavior detection. These patterns can be extracted based on the velocity of the tracklets of selected interest points. We first segment the foreground region based on motion features and extract interest points only in the foreground region. The velocities of the individual tracklets are used for estimating collectiveness, conflict, and mean speed. An SVM classifier is applied to detect violence/fight based on these features.

Finally, the results of the two SVM classifiers are combined at the score level to derive an overall confidence measure (between 0 and 1) of abnormal crowd event.
\end{subsection}

\end{section}
%~~~~~~~~~~~~~~~~~~~~~~~~~~~~~~~~~~~~~~~~~~~~~~~~~~~~~~~~~~~~~~
\begin{section}{Results}
%~~~~~~~~~~~~~~~~~~~~
\begin{subsection}{Image annotation}
We  annotated  images  from $3$ urban train stations,  indoor and outdoor, at $4$  locations (platform, gateway, lobby, escalator) during  periods of heavy crowds on $2$  weekdays; one day includes a train incident during which the entire platform is fully crowded, for  $420$ minutes of video overall.

For each frame, we annotate  each full body, head $\&$ shoulders, and head, occupying at least $30$ pixels on the cameras. We observed that below that threshold, human annotation was not reliable. The native image resolution is $704 \times 576$. We eventually create $63,114$ boxes over $4,200$ frames for low crowd scenarios, and $202,283$ boxes over $6,527$ frames for high crowd scenarios.
\end{subsection}
%~~~~~~~~~~~~~~~~~~~~
\begin{subsection}{Crowd flow and density estimation}
We use $90\%$ of the data for training and $10\%$ for testing. Results are provided in Table~\ref{table:countResults}.
% \begin{table}[htb!]
% \centering
% \begin{tabular}{cllllllll}
% \toprule
% Ground truth & \multicolumn{5}{c}{Low crowd}& \multicolumn{2}{c}{High crowd}\\
% $\#$ Frames & $215$ & $229$ & $173$ & $120$ & $26$ & $294$ & $47$\\
% People count & $[0,5]$ & $[6,10]$ & $[11,15]$ & $[16,20]$ & $[21,25]$ & $[26,50]$ & $[51,75]$\\
% \midrule
% Absolute error & \multicolumn{7}{c}{Occurrences per crowd level ($\%$)} \\
% $[0,5]$ & $80$ ($100$) & $89$ ($100$) & $74$ ($100$) & $70$ ($97$) & $54$ ($100$) & $71$ ($92$) & $61$ ($91$) \\
% $[6,10]$ & $19$ ($0$) & $11$ ($0$) & $25$ ($0$) & $28$ ($3$) & $19$ ($0$) & $22$ ($8$) & $26$ ($9$) \\
% $[11,15]$ & $1$ ($0$) & $0$ ($0$) & $1$ ($0$) & $2$ ($0$) & $27$ ($0$) & $7$ ($0$) & $9$ ($0$) \\
% $[16,20]$ & $0$ ($0$) & $0$ ($0$) & $0$ ($0$) & $0$ ($0$) & $0$ ($0$) & $0$ ($0$) & $4$ ($0$) \\
% \bottomrule
% \end{tabular}
% \caption{Confusion matrix for absolute error per crowd level. The people detection model is used for low crowd, while the regression-based counting is used for high crowd. Results from the fusion model are indicated in parenthesis.}
% \label{table:countResults}
% \end{table}
\begin{table}[htb!]
\centering
\begin{tabular}{ccccccccc}
\toprule
Ground truth & \multicolumn{4}{c}{Low crowd}& \multicolumn{4}{c}{High crowd}\\
%$\#$ Frames & $215$ & $229$ & $173$ & $120$ & $26$ & $230$ & $294$ & $47$ \\
People count & \multicolumn{2}{c}{$[0,10]$} & \multicolumn{2}{c}{$[11,25]$} & \multicolumn{2}{c}{$[26,50]$} & \multicolumn{2}{c}{$[51,75]$}\\
\midrule
Absolute error & \multicolumn{8}{c}{Occurrences per crowd level ($\%$)} \\
$[0,5]$ & $85$ & $\textbf{100}$ & $71$ & $\textbf{99}$ & $71$ & $\textbf{92}$ & $61$ & $\textbf{91}$ \\
$[6,10]$ & $15$ & $\textbf{0}$ & $26$ & $\textbf{1}$ & $22$ & $\textbf{8}$ & $26$ & $\textbf{9}$ \\
$[11,15]$ & $0$ & $\textbf{0}$ & $3$ & $\textbf{0}$ & $7$ & $\textbf{0}$ & $9$ & $\textbf{0}$ \\
$[16,20]$ & $0$ & $\textbf{0}$ & $0$ & $\textbf{0}$ & $0$ & $\textbf{0}$ & $4$ & $\textbf{0}$ \\
\bottomrule
\end{tabular}
\caption{Confusion matrix for absolute error per crowd level. The people detection model is used for low crowd, while the regression-based counting is used for high crowd. Results from the fusion model are indicated in bold.}
\label{table:countResults}
\end{table}

While the error increases with the crowd level, the fusion model significantly improves upon the base models. These results also improve on the performance from off-the-shelf software that we tested, by $10\%$ to $20\%$. In our applications, users were typically interested in count estimates within $5$ persons of the ground-truth counts for low crowds, and within $25\%$ of the ground-truth  for dense crowds. The work presented in this article demonstrated the feasibility of estimating crowd level with high accuracy in practical urban monitoring contexts.
\end{subsection}
%~~~~~~~~~~~~~~~~~~~~
 \begin{subsection}{Crowd Incident Detection}
The incident detection module was evaluated on the Violent Flows dataset, see~\cite{hassner2012violent}, which consists of 246 videos from YouTube. We followed the same protocol (with five-fold cross validation) as~\cite{hassner2012violent} for benchmarking. By varying the threshold for confidence value, the area under the receiver operating characteristic curve (AUC) of our system was found to be $87.9$, improving on the $85$ result from~\cite{hassner2012violent}.

Since the Violent Flows database focuses only on crowd violence, we also collected a small test set of $10$ videos, which were independently downloaded from YouTube. Among these $10$ videos, $7$ samples depict fights among individuals or small groups, while the remaining $3$ samples depict other activities with rapid motion such as group dancing or playing games.

In all the $7$ fight samples, our method detected an incident with a confidence value of greater than $0.4$. For the remaining $3$ samples, the confidence value of an  incident did not exceed $0.2$. The average time lag between the outbreak of violence and its detection was found to be $10$ seconds.
\end{subsection}
%~~~~~~~~~~~~~~~~~~~~
\end{section}

\section{Conclusion}
This short article presented an overview of a set of video-based techniques designed for real-world crowd insights and the management of dense crowds in urban environments, and described results from actual deployments.

%~~~~~~~~~~~~~~~~~~~~~~~~~~~~~~~~~~~~~~~~~~~~~~~~~~~~~~~~~~~~~~
% \bibliography{../../bib_roma}
% \bibliographystyle{plain}

\end{document}